\documentclass[conference]{IEEEtran}
\IEEEoverridecommandlockouts
\usepackage{cite}
\usepackage{amsmath,amssymb,amsfonts}
\usepackage{algorithmic}
\usepackage{graphicx}
\usepackage{textcomp}
\usepackage{multirow}
\usepackage[ruled,vlined]{algorithm2e}
\usepackage{xcolor}
\usepackage{framed}
\def\BibTeX{{\rm B\kern-.05em{\sc i\kern-.025em b}\kern-.08em
    T\kern-.1667em\lower.7ex\hbox{E}\kern-.125emX}}
\begin{document}

\title{Tweet to News Conversion: An Investigation into Unsupervised Controllable Text Generation\\}


\author{\IEEEauthorblockN{Zishan Ahmad\IEEEauthorrefmark{1}\textsuperscript{\textsection}, Mukuntha N S\IEEEauthorrefmark{2}\textsuperscript{\textsection}, Asif Ekbal\IEEEauthorrefmark{3} and
Pushpak Bhattacharyya\IEEEauthorrefmark{5}}
\IEEEauthorblockA{Department of Computer Science and Engineering,
Indian Institure of Technology Patna\\
 Patna, India\\
Email: \IEEEauthorrefmark{1}1821cs18@iitp.ac.in,
\IEEEauthorrefmark{2}mukuntha.cs16@iitp.ac.in,
\IEEEauthorrefmark{3}asif@iitp.ac.in,
\IEEEauthorrefmark{5}pb@iitp.ac.in}}
\maketitle
\begingroup\renewcommand\thefootnote{\textsection}
\footnotetext{equal contribution}
\endgroup
\maketitle

\begin{abstract}
Text generator systems have become extremely popular with the advent of recent deep learning models such as encoder-decoder. Controlling the information and style of the generated output without supervision is an important and challenging Natural Language Processing (NLP) task. In this paper, we define the task of constructing a coherent paragraph from a set of disaster domain tweets, without any parallel data. We tackle the problem by building two systems in pipeline. The first system focuses on unsupervised style transfer and converts the individual tweets into news sentences. The second system stitches together the outputs from the first system to form a coherent news paragraph.
We also propose a novel training mechanism, by splitting the sentences into propositions and training the second system to merge the sentences.
We create a validation and test set consisting of tweet-sets and their equivalent news paragraphs to perform empirical evaluation. 
In a completely unsupervised setting, our model was able to achieve a BLEU score of 19.32, while successfully transferring styles and joining tweets to form a meaningful news paragraph.
\end{abstract}


\section{Introduction}
\label{sec:intro}
Text generation using neural networks has become immensely popular in recent times due to the advent of sequence-to-sequence (seq2seq) \cite{kalchbrenner2013recurrent, cho2014properties} models. Long Short Term Memory (LSTM) \cite{sut2014seq} and modern Transformer \cite{vas2017atten} based models have been revolutionary for several tasks like machine translation, abstractive text summarization etc. However, the training of such models requires a huge amount of data, which is expensive and cumbersome to create. Although there are available large datasets for a few text generation tasks, creation of a dataset for every new task is not always practical or feasible.
To get around this issue, unsupervised text generation is increasingly being used, particularly in the area of Machine Translation \cite{atret2018unsup} and Style Transfer \cite{zhang2018style}.

Another problem with text generation is the control of information being generated. Models like GPT-2 \cite{gpt2} produces long paragraphs resembling human-written text, that are fluent to read. However, the input to the model is only the starting few words of the sequence\footnote{https://talktotransformer.com/}. The generated paragraph-although fluent, 
does not necessarily represent the information that we may want to convey. Thus, guiding the text generator to generate the information that is desired, and in the style that it is desired in, is a crucial task in text generation. Doing so in an unsupervised setting is a non trivial task, that poses the following challenges for the system.
\begin{itemize}
    \item The system should be able to achieve a good mapping between the input style and the desired style in the output without any supervision.
    \item The system should not generate extra information or miss some information that is provided at the input.
    \item The system should learn to stitch together the nuggets of information provided at the input, and to produce a coherent paragraph in the desired style. This too should be achieved without any parallel data.
\end{itemize}

In this paper we define a new task of converting a set of tweets into a news paragraph to explore the above challenges. The task not only poses a great challenge to the field of Machine Learning and Natural Language Processing, but is also an important task with respect to the applications. The system can be used to automatically generate news from the live tweet-stream. This can be crucial in emergency situations like disasters, for gaining comprehensive live situational data, in a non-noisy, readable, formal text format. Since we do not have any parallel data for the task, we have to develop unsupervised techniques to solve this problem. We try to tackle the problem by dividing it into two parts, \textbf{i). Part 1:} Converting individual tweets to formal news text format, and \textbf{ii). Part 2:} Stitching together the converted tweets to form a comprehensive and fluent news paragraph that contains all the information provided in the tweets.

We build two separate systems to tackle Part 1 and Part 2. The entire system is structured in a pipeline where Part 1 would first take the input tweets, and then convert them into news format. The second part (i.e. Part 2) would then take the converted tweets from the first part and output a coherent news paragraph that contains information from all the tweets. For Part 1, we model the task as an unsupervised style transfer task of transferring from tweet-style to news-style. Although unsupervised style transfer has been explored before, it has mostly been carried out to change the sentiment of a text \cite{zhang2018style,yang2018unsup}. Li et al. \cite{li2018style} showed that sentiment style transfer can be achieved by only changing a few attribute markers (such as `good' to `bad'), while leaving the sentences largely unchanged. Thus, it is easier to capture sentiment pattern in text than the pattern that defines informal (tweet) or formal text (news). Such a task is more complex, requiring injecting articles wherever necessary, replacing informal words with formal words, discarding hashtag words where it is not important, retaining the hashtags wherever important and getting rid of other noise in the text as well, for which rules cannot be written.

The second part (i.e. Part 2) is tackled by a separate system that takes the outputs generated by the first part and joins them together to generate a paragraph. The task to be tackled in Part 2 is again non trivial, as it has to get rid of redundancies in information in different tweets, and keep the unique information. While doing so, it also has to figure out a way of joining sentences wherever required and not joining sentences when not required. In this paper, we develop a system that only learns to join sentences together. 
We propose a novel unsupervised approach to achieve joining of separate sentences, where we use news sentences that have been broken down into propositions.
After obtaining news sentences and their clauses, we train an encoder-decoder based model that learns to stitch these propositions together to form a coherent sentence. We also create and release a validation and test set for the task to quantitatively measure the system's performance.

Our evaluation shows that our method performs well in terms of both automatic and human evaluation metrics.
In summary, our current work contributes:
\begin{itemize}
\item A new method to convert informal and noisy tweet content into a more formal news-text format.
\item A novel training regime to stitch together the information nuggets into long coherent sentences.
\item An evaluation dataset for the task, consisting of 1265 instances of a set of four different domains of disasters 
and their corresponding news paragraphs.
\end{itemize}

\subsection{Problem Definition and Motivation}
\label{subsec:prob}
Given a set of four tweets of a particular event from the disaster domain, the task is to form a paragraph in news format that contains all the information provided in the tweets and does not produce extra information. The model should thus form a formal and comprehensible news paragraph from the given noisy input (i.e. tweets). Following is an example of the task:
\begin{itemize}
    \item \textbf{Tweet 1 (Input 1):} breaking: at least 126 killed in taliban attack on pakistani school.
    \item \textbf{Tweet 2 (Input 2):} update: \#peshawarattack over, all hostage-takers dead - police.
    \item \textbf{Tweet 3 (Input 3):} dozens of children killed as taliban gunmen storm peshawar school.
    \item \textbf{Tweet 4 (Input 4):} \#pakistan school attack over, all six attackers are dead. \#peshawarattack \#talibanattacksschool
    \item \textbf{News (Output):} Taliban gunmen stormed a school in Pakistan, killing at least 126. Police stated that the Peshawar attack is over as all six hostage-takers are dead.
\end{itemize}
The above problem is defined to tackle the challenges mentioned in Section \ref{sec:intro}. The difficulty of this task is also exacerbated as tweets are extremely noisy input source.

This task is motivated to address the important Natural Language Processing (NLP) challenges such as information guided and style controlled unsupervised text generation. Also, joining pieces of information to generate a well formed paragraph has been largely unexplored. The system built to solve this task has useful application as an automated news generator, that takes live tweets as inputs and generates news paragraph. This is useful especially in disaster domain, as it can provide better situational awareness. The users will not have to explore endless noisy tweets but can get one news paragraph that provides live information in a clear news format. This system, thus, helps us take another step in automated journalism.

\section{Related Work}
\label{sec:ref}
The task defined in this paper has not been previously tackled, to our knowledge. However, there are a few tasks that aim to solve a few sub-tasks and adjacent tasks. One of the sub-tasks that has previously been explored is `style-transfer in text'.  Nahas et al. \cite{nahas2019unsup} used LSTM based attentive encoder-decoder network to style transfer between old and modern Turkish text. They had parallel corpus between old and modern Turkish language to train their model. Since it's often not possible to obtain parallel corpora, semi-supervised and unsupervised text style transfer are also very actively being explored 
in the machine learning community. Shang et al. \cite{shang2019semi} proposed a semi-supervised method of transferring text styles between Chinese ancient poetry style and modern Chinese style. They also tested their experiment on style transfer between formal and informal English text. 

Unsupervised style transfer is often achieved by using unsupervised neural machine translation techniques \cite{zhang2018style}, such as back-translating data in one style to create synthetic parallel data. Jin et al. \cite{jin2019unsup} introduced iterative matching of synthetic data produced by unsupervised Neural Machine Translation (NMT) technique. They used cosine similarity and word mover distance to keep the best synthetic parallel data. They showed that their method outperforms unsupervised NMT method in sentiment transfer and formality transfer tasks. Yang et al. used \cite{yang2018unsup} language models to discriminate between the sentiment of the generated text by unsupervised NMT method. This way they were able to propagate error at every step of generation and achieve better performance in transferring sentiment of the text. Luo et al. \cite{luo2019dual} used dual reinforcement learning to better capture style and content and improve the style transfer. They evaluated their model for sentiment and formality transfer tasks.

Another task that may seem similar to our task is text summarization. Traditional text summarization tasks aim to condense long text into a small summary. This summary may even exclude some information. In our task, input information should not be lost at the output, even though redundancy has to be removed. Also the length of the output text may or may not be shorter than the input text. Abstractive summarization tasks require a huge amount of parallel data. Various tasks like headline generation \cite{lit2019head, li2017rec}, wikipedia summarization \cite{liu2018gen} and news document summarization \cite{tan2017abs} have been tackled using supervised abstractive text summarization. 
Encoder-decoder neural models are the most popular techniques for abstractive summarization. Some works like the one presented in Liu et al. \cite{liu2018gen} use discriminator to identify human and machine created text, to improve the performance of their encoder-decoder model. 

Unsupervised summarization is usually achieved through extractive summarization techniques. Handcrafted features are often used \cite{kiy2011opt,chen2002sum,san2014sum} to extract the sentences. These extracted sentences are assumed to represent the summary of the text. 
Shen et al. \cite{shen2007sum} used trained a Conditional Random Field (CRF) classifier to label sentences as part of summary or not the part of summary. Although extractive summary, this method still needed labelled data. 

The advantage of extractive text summarization is that it can be achieved in unsupervised manner. However, a major drawback with extractive text summarization techniques is that the summaries formed are not coherent and are disjoint. The summary formed is not a fluent paragraph, but a collection of sentences that represents different information. Abstractive summarization forms well formed, comprehensible summaries, however requires a huge amount of parallel data. In our knowledge there has been no prior work that performed 
unsupervised abstractive text summarization. Although the work in this paper is different from summarization, it still tackles the problem of removing redundancies and joining sentences to form a coherent paragraph.

\section{Proposed Methodology}
\label{sec:method}
As discussed in Section \ref{sec:intro}, we solve the problem in two parts and build separate system for each part. In part 1, we model the problem as a style transfer task from tweets to news, whereas in part 2, the output from part 1 is taken and stitched together to form a paragraph. Throughout our experiments, we use a transformer model \cite{vas2017atten} as our encoder-decoder model, and initialize it with the weights of cross-lingual language model (XLM) \cite{lample2019cross}. The details of each system is discussed in this section.
\subsection{Cross-lingual language model (XLM)}
\label{subsec:p1}
Lample et al. \cite{lample2019cross} proposed XLM to create a transformer \cite{vas2017atten} based cross-lingual language model. To achieve this, three different training mechanisms are followed:
\begin{enumerate}
    \item \textbf{Causal Language Modeling (CLM):} CLM is modelled to predict the probability of the next word given the previous context in the sentence $P(w_t|w_1,w_2,...w_{t-1},\theta)$.
    \item \textbf{Masked Language Modeling (MLM):} MLM \cite{dev2018bert} is achieved by randomly sampling 15\% of the input BPE (byte-pair encoded) tokens. These sampled tokens are replaced by: (i). [MASK] token 80\% of the time, (ii). Random token 10\% of the time and (iii). Are left unchanged 10\% of the time.
    \item \textbf{Translation Language Modeling (TLM):} The TLM is modelled as an extension of MLM. Here, instead of using only monolingual text-streams, parallel sentences of different languages are concatenated. Random masking is then performed in all the sentences. To predict a  masked word in one language, the model can either look at the surrounding context of the same language sentence, or it can look into parallel sentences of other language to get hint of the mask word. This is the way how the model learns to find alignment between the different languages.
\end{enumerate}
The XLM model uses only one encoder and one decoder with the language embeddings of several languages, instead of using separate encoders and decoders to create a multi-lingual system. Thus, this model consumes less memory than multi-encoder-decoder systems. In our experiments, since we do not have any parallel data, we only use MLM. To use this model, we treated tweets and news as two styles, and train separate style embeddings for them, analogous to the language embeddings used in XLM.

\subsection{Proposed Model-First Part }
\label{sec:p1}
To solve the style transfer task in the first part (i.e. Part 1), two different systems are built. We take two non-parallel datasets of two different styles, $X = \{x_1, x_2 \ldots x_m\}$ consisting of tweets and $Y = \{y_1, y_2 \ldots y_n\}$ consisting of news sentences. Let $l_1$ denotes the tweet-style and $l_2$ denotes the news-style. Our goal is to model the conditional distributions $p(x|y)$ and $p(y|x)$; i.e., to transfer data $x$ of the tweet style $l_{1}$ to the news style $l_{2}$. The encoder $E$ encodes inputs $x$ and $y$ to give content vectors $z_x = E(x, l_{1})$ and $z_y = E(y, l_{2})$. The decoder $D$ decodes $z_x$ and $z_y$ to give $\hat{x} = D(z, l_{2})$ and $\hat{y} = (z, l_{1})$ respectively, where $z$ is any content vector outputted by the encoder. Also, let $\theta_{E}$ and $\theta_{D}$ represent the parameters of the encoder and decoder, respectively.

\subsubsection{XLM-STY}
\label{subsub:xmlsty}
We use unsupervised neural machine translation (UNMT) steps \cite{artetxe2018unsupMT} to achieve style transfer between tweets and news. The steps followed in this model are listed below:
\begin{itemize}
    \item \textit{De-noising step:} In this step noise in the form of random masking, shuffling and dropping of BPE tokens is added to the encoder input sentence, while the decoder is trained to reconstruct the original de-noised sentence. This is done in order to make the model learn the two style distributions. The reconstruction loss $L_{rec}$ is given in Equation \eqref{eq:l_rec}. Here, $x \in X$ is a tweet, and $y \in Y$ is a news sentence. $\hat{x}_{dn} ~ D(E(C(x), l1), l1)$ denotes that $\hat{x}_{dn}$ is a reconstruction of the noised version $C(x)$ of the input tweet $x$, where $C$ is a function that adds random noise to $x$.
    \begin{equation}
        \begin{split}
            L_{rec} (\theta_{E}, \theta_{D}) = \mathbb{E}_{x, \hat{x}_{dn} \sim D(E(C(x), l_{1}), l_{1})}(-log(p(\hat{x}_{dn}|x))) \\
    		+ \mathbb{E}_{y, \hat{y}_{dn} \sim D(E(C(y), l_{2}), l_{2})}(-log(p(\hat{y}_{dn}|y)))
        \end{split}
        \label{eq:l_rec}
    \end{equation}
    \item \textit{On-the-fly-back-translation:}
    While the de-noising step helps learn the individual style distributions, it does not help learn the transfer of a sentence from one style to the other. So in this step, we use our current style-transfer model $M_{12}$ to a news sentence $x$ to obtain a synthetic version $y' = M_{12}(x)$ in the target tweet domain. These are then used as parallel data to help train the encoder and the decoder in the tweet-to-news direction. The same is repeated with a tweet $y$ and a generated synthetic news sentence $x' = M_{21}(y)$ to train the model in the news-to-tweet direction. Here, the style-transfer models from which synthetic outputs are sampled, are given by $M_{12}(x) = D(E(x, l_1), l_2)$ and $M_{21}(y) = E(D(y, l_2), l_1)$. Equation \eqref{eq:l_bt} defines the back-translation loss $L_{bt}$.
    \begin{equation}
        \begin{split}
            L_{bt} (\theta_{E}, \theta_{D}) = \mathbb{E}_{x, \hat{x}_{bt} \sim D(E(M_{12}(x), l_{2}), l_{1})}(-log(p(\hat{x}_{bt}|x))) \\
		+ \mathbb{E}_{y, \hat{y}_{bt} \sim D(E(M_{21}(y), l_{2}), l_{2})}(-log(p(\hat{y}_{bt}|y)))
        \end{split}
        \label{eq:l_bt}
    \end{equation}
    
\end{itemize}

\subsubsection{XLM-STY-DIS + SYN}
\label{subsub:xlmstydisc}
\begin{figure}
    \centering
    \includegraphics[width=0.45\textwidth]{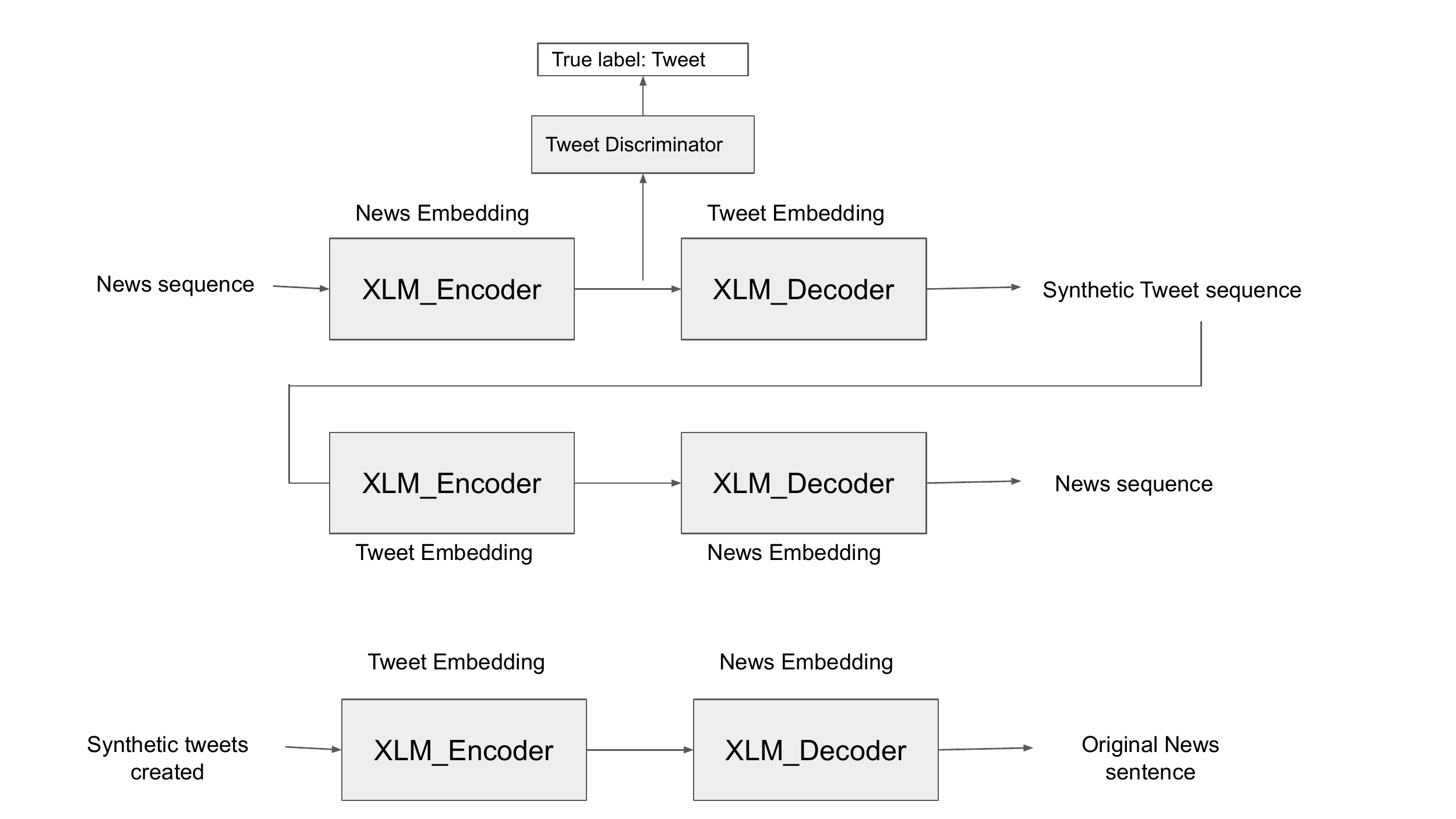}
    \caption{Block diagram of the `XLM-STY-DIS + SYN' architecture}
    \label{fig:xlm-part1}
\end{figure}

This model introduces two new modifications to the baseline \textit{XLM-STY} model (c.f. Figure \ref{fig:xlm-part1}):
\begin{itemize}
    \item \textit{Adversarial training (DIS):} We additionally train a Gated Recurrent Unit (GRU) based discriminator $D$ that takes in the content vectors $z_x = E(x, l_{1})$ and $z_y = E(y, l_{2})$ produced by the encoder and classify them as tweet or news. This is done in order to obtain a style-invariant representation at the encoder's output, thus aligning the two $z$ distributions. This style invariant content representation is then fed to the discriminator, which decodes the representation into the desired style. The discriminator is trained as a binary classifier that outputs the probability $p(l_1 | z)$ that a given latent content vector $z$ comes from a tweet, with $p(l_2 | z) = 1 - p(l_1 | z)$ being the probability that it comes from a news sentence. The adversarial loss $L_D$ used to train the discriminator weights $\theta_{D}$ is given by the following cross-entropy loss:
    \begin{equation}
    \begin{split}
        L_{D}(\theta_{D}|\theta_{E}) = \mathbb{E}_{x \sim X, z_x \sim E(x, l_1)}(- log(p(l_1 | z_x))) \\ + \mathbb{E}_{y \sim Y, z_y \sim E(y, l_2)}(- log (1 - p(l_1 | z_y)))
    \end{split}
    \end{equation}
    The encoder is trained with the reverse objective:
    \begin{equation}
        L_{adv} = - L_D
    \end{equation}
    
    \item \textit{Synthetic parallel-data training step (SYN):} To make the model more robust, we add another step to the training process. Apart from \textit{UNMT} and \textit{Adversarial Training} step, the encoder-decoder model is also trained with a synthetic tweet data that we create. Tweets have some unique properties \cite{kauf2010syn}, like random spelling mistakes, random hashtags, hashtags that convey important information etc. These properties cannot be induced by our synthetic tweet generator, as the generator is unlikely to produce random spelling mistakes, and also cannot generate random hashtags. We try to mimic these properties by creating a synthetic parallel tweet-data using news sentences and following the steps below:
    \begin{itemize}
        \item \textbf{Step 1:} Paraphrase the English news sentence by using pre-trained translation modules to translate it to a distant language and translating it back to English. This way the paraphrasing is a little noisy, which is helpful for our task.
        \item \textbf{Step 2:} Introduce random spelling mistakes like flipping random characters, dropping random characters and dropping all the vowels from the word with a probability of 15\%.
        \item \textbf{Step 3:} Randomly hashtag named entities from the sentences with a probability of 15\%. Also we have a list of random hashtags used from the tweet corpus. Using this list we randomly inject a hashtag into each tweet with a 15\% probability.
    \end{itemize}
    
    This synthetic data is used to train the same encoder-decoder model in a separate step. The reconstruction loss $L_{rec}$ used for this training is given in Equation \ref{eq:l_rec_syn}. Here, $H(x)$ is the function that converts the news sentences into synthetic tweets by following the aforementioned steps.
    \begin{equation}
    \label{eq:l_rec_syn}
        L_{syn} (\theta_{E}, \theta_{D}) = \mathbb{E}_{y, \hat{y}_{syn} \sim D(E(H(y), l_{2}), l_{1})}-log(p(\hat{y}_{syn}|y))
    \end{equation}
\end{itemize}

{}

\subsection{Proposed Model-Second Part}
\label{sec:p2}

To tackle Part 2, we train XLM-MERGE, a model capable of stitching together the information nuggets from the news-style sentences obtained from the first part, i.e. Part 1. To achieve this, we break the news sentences into propositions using ClauseIE \cite{corro2013clausie}. ClauseIE is a clause-based framework to open information extraction from sentences. It is based on dependency parse trees, and extracts complex propositions from the news sentences. Given a news sentence $y$, ClauseIE extracts a set of `propositions' ${p_1, p_2, \ldots p_n}$. Each proposition extracted consists of a subject phrase (for example ``a young woman"), a relation phrase (``has been airlifted") and zero, one or more argument phrases (``to hospital"). We join these propositions to form sentences (``a young woman has been airlifted to hospital"). We then train an encoder-decoder architecture to merge these propositions back into their original sentence with the following cross-entropy loss $L_m$.
\begin{equation}
    L_{m} (\theta_{E}, \theta_{D}) = \mathbb{E}_{y, \hat{y}_{m} \sim D(E(P(y), l_{1}), l_{2})}(-log(p(\hat{y}_{m}|y)))
\end{equation}
Here, $P(y) = p_1 \Vert p_2 \Vert \ldots \Vert p_n$, a paragraph of appended proposition sentences extracted from the news sentence $y$. $\hat{y}_{m}$ is a reconstruction of the original sentence from $P(y)$.
We initialize both encoder and decoder with the pretrained XLM Masked Language Model. Let $\theta_{E}$ and $\theta_{D}$ be the parameters of the encoder and decoder, respectively, and $l_1$ and $l_2$ be the language-embeddings used for propositions and full sentences, respectively.

The following is an example of propositions extracted by ClauseIE:
\begin{itemize}
    \item \textbf{Sentence}: a young woman has been airlifted to hospital after her car veered into trees in the la trobe valley.
    \item \textbf{List of Propositions Extracted}: a young woman has been airlifted to hospital. a young woman has been airlifted after her car veered into trees in the la trobe valley. a young woman has been airlifted into trees in the la trobe valley. a young woman has been airlifted in the la trobe valley. she has car. her car veered after her car veered into trees in the la trobe valley. her car veered into trees in the la trobe valley. her car veered in the la trobe valley 
\end{itemize}

As we can see from the above example, the propositions contain redundant information, and are sometimes noisy. This is very similar to tweets. While a majority of the propositions are useful and help us regenerate the sentence, we also find that a few of them can be incorrect (``a young woman has been airlifted into trees in the la trobe valley."). However, this noise is only present at the source-side and makes the model more robust to noisy data coming from the first part (i.e. Part 1). Unsupervised machine translation methods often employ back-translation to exploit the error-free data on the target side, leading to more fluent outputs. Similarly, we exploit the error-free data on the target-side, here - the original news sentences.

\section{Datasets and Experiments}
\label{sec:exp}
Although the system we build is completely unsupervised, we create a validation and test set for evaluating the performance of our model. We also use non-parallel news and tweet data. The detailed description and statistics of the datasets used is mentioned in this section. We conduct several experiments and perform ablation tests to highlight the importance of each module. In this section, we list the different experiments done and the experimental setups for all the experiments.

\begin{table}[t]
\centering
\caption{Data distribution}
\label{tab:data}
\begin{tabular}{|l|l|}\hline
\textbf{Dataset} & \textbf{Number of Instances} \\\hline
News-Tweet-Parallel Validation & 265 \\
News-Tweet-Parallel Test & 1,000 \\
\hline
Tweet-Non-Parallel & 45,295 \\
WMT-Non-Parallel & 171,400 \\
\hline
News-Clause-Pair & 126,120\\
\hline
\end{tabular}
\end{table}

\subsection{Datasets}
\label{subsec:data}
To create the test and validation sets, we crawl disaster domain tweets using hashtags for different disaster events. We build a disaster domain classifier based on the work of Nguyen et al. \cite{ngu2017rob}. This classifier was used to filter out the tweets that were relevant to the disaster domain. K-means clustering was used to cluster the tweets into four different topics. Clustering ensures that similar tweets are bunched together. We select one tweet from each of the clusters and give it to the human annotators to create a news paragraph such that, it represents all the information in the tweet. Selection is done from the different clusters to ensure diversity in tweets and avoid redundancy. Three annotators having graduate-level expertise in English language are used to create this dataset. The statistics of the curated dataset is given in Table \ref{tab:data} (News-Tweet-Parallel Validation and Test).

For training the unsupervised model, non-parallel tweet and news data were used. We crawl 22 million open domain tweets, and obtain the news sentences from the WMT-2017 dataset. We use our domain classifier to filter out the disaster domain tweets from the open domain tweets. To obtain similar, in-domain news sentences from the WMT-2017 dataset, we use TF-IDF and cosine similarity between the filtered tweets and the news sentences. The data distribution for these non-parallel data is shown in Table \ref{tab:data} (Tweet-Non-Parallel and WMT-Non-Parallel).

The dataset we prepare to train XLM-MERGE was prepared by extracting propositions using ClauseIE \cite{corro2013clausie} from these disaster domain news sentences. We drop all the sentences with fewer than two propositions. The concatenation of the propositions is limited to 512 BPE tokens, and longer sequences are truncated. The details of this dataset is also mentioned in Table \ref{tab:data} (News-Clause-Pair).

\subsection{Experiments Conducted}
\label{subsec:exps}
We conduct the following experiments in this paper to better understand the significance of each step in the training process.
\begin{itemize}
    \item \textbf{XLM-STY:} Since there are no previous models specific to this task, we use traditional unsupervised style transfer, using XLM as our baseline experiment (discussed in Section \ref{subsub:xmlsty}). In this experiment, we use the XLM-STY model and exclude the part 2 system i.e. XLM-MERGE. During evaluation, we pass four tweets concatenated together to the system and a news-style paragraph is thus obtained at the output.
    \item \textbf{XLM-STY-DIS + SYN:} In this experiment we use the XLM-STY-DIS model along with the synthetic-parallel-tweets training step. The XLM-MERGE model is not used here. We then pass the 4 tweets separately to the XLM-STY-DIS model and the outputs from XLM-STY-DIS model are concatenated to form the paragraph.
    \item \textbf{XLM-MERGE} In this experiment, four tweets are directly merged to form a paragraph using the XLM-MERGE model. There was no style transfer applied to the tweets.
    \item \textbf{XLM-STY + SYN + XLM-MERGE:} In this experiment, after obtaining the results from XLM-STY, XLM-MERGE is used to merge the 4 output sentences into a paragraph. We do not use a discriminator in training XLM-STY, but use the synthetic-parallel-tweets training step.
    \item \textbf{XLM-STY-DIS + XLM-MERGE:} In this experiment the result obtained from the XLM-STY-DIS experiment is merged into a paragraph using the XLM-MERGE model.
    \item \textbf{XLM-STY-DIS + SYN + XLM-MERGE:} In this experiment the outputs obtained from XLM-STY-DIS + SYN experiment is merged to form a paragraph using the XLM-MERGE model.
\end{itemize}

\subsection{Experimental Setup}
\label{subsec:exp}
\begin{figure}[hbt!]
 \centering
 \label{fig:xlm-mlm}
    \includegraphics[width=0.45\textwidth]{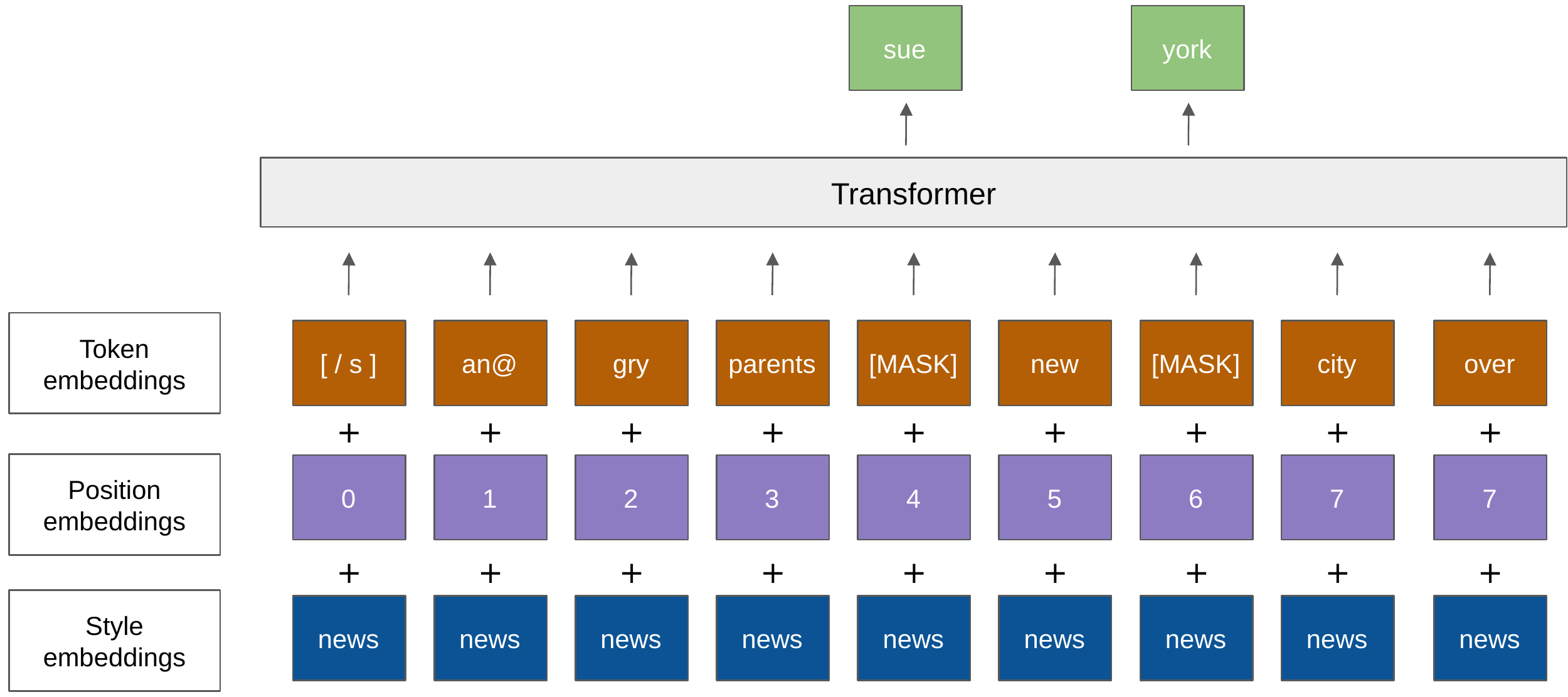}
    \caption{XLM Masked Language Model (MLM) fine-tuning}
 \end{figure}

Lample et al. (2019) \cite{lample2019cross} demonstrated that initializing the encoder and decoder of a transformer network with their pre-trained model can significantly improve the results for unsupervised machine translation. Since our task is to transfer style between news and tweets, we fine-tune an XLM-based Masked Language Model (MLM) with news and tweets as the two styles. For our initialization, we use the 15-language XLM-MLM model released by Lample et al. (2019) \cite{lample2019cross}. As mentioned in Section \ref{subsec:p1}, XLM uses only one shared encoder and one shared decoder for all the languages, while using different language-specific embeddings that get added to each token's embedding at the input. We initialize all of our model's parameters from theirs, and initialize the language-specific embeddings for news and tweets with the embedding corresponding to the English language from XLM. We then fine-tune our MLM model on the same Cloze task \cite{clozetask}, as shown in \ref{fig:xlm-mlm}. Following Lample et al., we randomly sample 15\% of the input BPE tokens and replace them by a [MASK] token 80\% of the time, by a random token 10\% of the time while otherwise keeping them unchanged. During training, we use streams of 256 tokens, and mini-batch sizes of 8.

In all our experiments for style-transfer and merging, we use a Transformer as our encoder-decoder model and initialize the encoders and decoders with our fine-tuned tweet-news XLM-MLM model. We use transformer models with 8 attention heads, 1024 hidden-units and GELU activations, and use 6-layers each for our encoders and decoders. We use the Adam optimizer with a learning rate of $1 \times 10^{-5}$, and mini-batches of size 4. Following Artetxe et al. (2018) \cite{artetxe2018unsupMT}, we back-translate each mini-batch on-the-fly using the model that is training itself. We also perform the adversarial-training and synthetic-parallel-data training (SYN) steps on-the-fly in the same manner. During training, we alternate between the training objectives - $L_{rec}$, $L_{bt}$, $L_D$, $L_{adv}$ and $L_{syn}$, with one-minibatch for each objective. For training XLM-STY and XLM-STY-DIS, we use the non-parallel WMT and Tweet datasets for training, and use the News-Tweet-Parallel data as validation and test. To train XLM-MERGE, we use the News-Clause-Pair dataset. During evaluation, if we pass all the four style-converted tweets to the XLM-MERGE model together, the model would output a single sentence with all the information merged into one sentence. To avoid this, we pass the style-converted tweets two at a time to the XLM-MERGE model for merging.

\section{Results and Analysis}
\label{sec:res}

\begin{table}[t]
\centering
\caption{Five-point scales and definition guidelines for human assessments on adequacy and fluency}
\label{tab:human}
\begin{tabular}{lll}\hline
\textbf{Metric} & \textbf{Score} & \textbf{Definition} \\\hline
\multirow{5}{*}{Adequacy} & 1 & None of the meaning is preserved \\
& 2 & Little of the meaning is preserved \\
& 3 & Much of the meaning is preserved \\
& 4 & Most of the meaning is preserved \\
& 5 & All the meaning is preserved \\
\hline
\multirow{5}{*}{Fluency} & 1 & Incomprehensible target sentence \\
& 2 & Dis-fluent target sentence \\
& 3 & Non-native kind of target sentence \\
& 4 & Good quality target sentence \\
& 5 & Flawless target sentence \\
\hline
\end{tabular}
\end{table}



\begin{table*}[t]
\centering
\caption{BLEU, adequacy and fluency  scores for the different experiments conducted}
\label{tab:result}
\begin{tabular}{ccccllll}\hline
\textbf{Experiments} & \textbf{DIS} & \textbf{SYN} & \textbf{MERGE} & \textbf{BLEU} &  \textbf{Adequacy} & \textbf{Fluency} \\\hline
XLM-STY & No & No & No & 14.34 & 2.206 & 2.245  \\
XLM-STY-DIS + SYN & Yes & Yes & No  & 18.28  & \textbf{3.52} & 2.34\\
XLM-MERGE  & No & No & Yes & 19.04 & 2.7275 & 2.63\\
XLM-STY-DIS + XLM-MERGE & Yes & No & Yes  & 16.31 & 2.74 & 2.7\\
XLM-STY + SYN + XLM-MERGE & No & Yes & Yes & 19.04 & 3.04 & 3.02\\
XLM-STY-DIS + SYN + XLM-MERGE & Yes & Yes & Yes & \textbf{19.32} & 3.04 & \textbf{3.2}\\
\hline

\end{tabular}
\end{table*}

To evaluate the results of our experiments, we use both automatic and manual evaluation metrics. For automatic evaluation we use the tokenized BLEU scores computed using multi-BLEU pearl script included in Moses \footnote{http://www.statmt.org/moses/}.

For manual evaluation we calculate adequacy and fluency of the generated text by following five-point scale system used by Rafael et al. \cite{banchs2015adequacy}. While adequacy is the measure of source information (meaning) preservation at the output, fluency measures the quality of the target language constructions used in the translation. The grading system used for rating the adequacy and fluency of the output is detailed in Table \ref{tab:human}. The human evaluation was done by two human annotators with graduate level exposure. We randomly sampled 50 outputs from each experiment (300 in total), and distributed it to the annotators. They were asked to follow the grading scheme in Table \ref{tab:human} to score the outputs. We compute inter-annotator consistency using Fleiss' \cite{fleiss1971measuring} kappa. We found the kappa of adequacy and fluency to be 0.74 and 0.76, respectively.

Results for the models in terms of BLEU scores are reported in Table \ref{tab:result}. Our baseline model `XLM-STY' that uses vanilla XLM for style transfer yields a BLEU score of 14.34. Our final model `XLM-STY-DIS + SYN + XLM-MERGE' outperforms the baseline in terms of BLEU (+4.98), adequacy (+0.83) and fluency (+0.96). The performance gains were found to be statistically significant \footnote{Stastical significance t-test \cite{welch1947generalization} was performed at 5\% significance level}. The model with best adequacy was found to be `XLM-STY-DIS + SYN'. It is because this model's output is the input to the `XLM-MERGE' model, therefore the error in terms of adequacy is propagated forward. Therefore, the full model cannot have a better adequacy than this model. Another interesting observation from the results is that using only `XLM-MERGE' on raw tweets produces good BLEU scores and fluency. However, the adequacy is low since the `XLM-MERGE' tries to merge all the information, and treats noise in tweets as important information too. 

The ablations clearly show the importance of each component of our model. In the experiment `XLM-STY + SYN + XLM-MERGE' after removing the discriminator from the final model we see a reduction in the BLEU score by 0.28, and the fluency by 0.18. In the experiment `XLM-STY-DIS + XLM-MERGE' the synthetic-parallel training step is removed, resulting in reduction in BLEU score by 0.28, adequacy by 0.30 and fluency by 0.5.

Some examples of the outputs produced by our model is shown below,

\begin{framed}
\noindent \textbf{Tweet1}:
rain brings relief to dhaka as cyclone faniapproaches . \\
\noindent \textbf{Tweet2}:
two people are reported to have died and more than a million have fled their homes after \#cyclonefani made landfall on india's east coast. more on this story here:.
\\ \noindent \textbf{Tweet3}:
thousands evacuated in eastern india as cyclone fani approaches.
\\ \noindent \textbf{Tweet4}:
'extremely severe' cyclone fani to hit south of puri on friday: ndma.
\\ \noindent \textbf{OUT}:
rain brings relief to dhaka as two people are reported to have died and more than a million have fled their homes after biclonefani made landfall on india's east coast. more on this story . in eastern india as cyclone approaches , 's tremely severe .
\end{framed}
\begin{framed}
\noindent \textbf{Tweet1}: deadly cyclone fani pummels india with wind gusts over 120 mph and flooding rain:.
\\ \noindent \textbf{Tweet2}: cyclone fani wreaks havoc across east coast of india leaving three dead  telegraph.
\\ \noindent \textbf{Tweet3}: 3m people to evacuate as bangladesh braces for cyclonefani.
\\ \noindent \textbf{Tweet4}: navys dornier aircrafts aerial photos show massive damage due to cyclone fani inpuri.
\noindent \\
\textbf{OUT}: deadly cyclone pummels india with wind gusts over 120mph and flooding rain , leaving three dead telegraph . as bangladesh braces for biclonefani , navys


\end{framed}

In the first example we can see that the model is able to merge the first two tweets fluently as they are both talking about relief and casualties. It makes a spelling mistake due to the noisy hash-tag. The third and fourth tweets contained more noise in the form of phrase `more story on this'. This confused the model and it skipped the information in that sentence. In the second example too we can see the first sentence, that is obtained by joining two tweets, is fluent and complete. It successfully merges the tweets and gets rid of the redundant information. However, the topic of the last two tweets were completely different. One tweet talks about evacuation and the other tweet is about aerial photographs. Such sentences cannot be joined, and hence our model fails to do so. Since we are randomly merging tweets, such issues persist. 

Below are some examples of the outputs of XLM-MERGE, from the test data we prepared for the clause-merging task. These outputs show that XLM-MERGE is sufficiently robust, and learns to produce succinct sentences when provided one or more clauses with redundant information. This reiterates that the outputs of our whole model can be improved if the style-transfer step produces less noisy outputs from the tweets.

\begin{framed}
\noindent
\textbf{IN}: telegraph media group helped fund. telegraph media group helped the project. telegraph media group owns the daily telegraph. the daily telegraph is daily. \\
\textbf{OUT}: telegraph media group, which owns the daily telegraph, helped fund the project.
\noindent \\ \\
\textbf{IN}: he will participate in seminars and on-campus activities. he will participate throughout the year.
\noindent \\
\textbf{OUT}: he will participate in seminars and on-campus activities throughout the year.

\end{framed}

\section{Conclusion and Future Work}
\label{sec:conc}

In this paper, we have explored a new task of creating a news paragraph from a given set of tweets in the disaster domain in an unsupervised manner. We intend this to be an initial step towards solving the problem of unsupervised guided text generation, where we intend to control the style and the content of generated text. We propose a pipeline system with two models to tackle this problem. Our first model transfers the style of informal and noisy tweets into more formal news-like sentences in an unsupervised setting. Our second model solves the problem of merging together several information nuggets across sentences to form a single paragraph. We also prepare and release an evaluation dataset for the task. We propose novel methods to train both our style transfer and merging models, and show through automated and human evaluations that our proposed methodology can outperform the baseline style transfer models.

In the future, we would like to explore better merging strategies for sentences, where the model learns when to merge and when to avoid merging sentences. We would also like to explore an end-to-end strategy to achieve merging and style transfer of sentences in one go, to avoid the propagation of error from .

\section*{Acknowledgment}
\label{sec:ack}
The research reported in this paper is an outcome of the project titled ``A Platform for Cross-lingual and Multi-lingual Event Monitoring in Indian Languages", supported by IMPRINT-1, MHRD, Govt. of India, and MeiTY, Govt. of India. Asif Ekbal acknowledges the Young Faculty Research Fellowship (YFRF), supported by Visvesvaraya PhD scheme for Electronics and IT, Ministry of Electronics and Information Technology (MeitY), Government of India, being implemented by Digital India Corporation (formerly
Media Lab Asia). 

\bibliographystyle{IEEEtran}
\bibliography{bibliography}

\begin{thebibliography}{10}
\providecommand{\url}[1]{#1}
\csname url@samestyle\endcsname
\providecommand{\newblock}{\relax}
\providecommand{\bibinfo}[2]{#2}
\providecommand{\BIBentrySTDinterwordspacing}{\spaceskip=0pt\relax}
\providecommand{\BIBentryALTinterwordstretchfactor}{4}
\providecommand{\BIBentryALTinterwordspacing}{\spaceskip=\fontdimen2\font plus
\BIBentryALTinterwordstretchfactor\fontdimen3\font minus
  \fontdimen4\font\relax}
\providecommand{\BIBforeignlanguage}[2]{{%
\expandafter\ifx\csname l@#1\endcsname\relax
\typeout{** WARNING: IEEEtran.bst: No hyphenation pattern has been}%
\typeout{** loaded for the language `#1'. Using the pattern for}%
\typeout{** the default language instead.}%
\else
\language=\csname l@#1\endcsname
\fi
#2}}
\providecommand{\BIBdecl}{\relax}
\BIBdecl

\bibitem{kalchbrenner2013recurrent}
N.~Kalchbrenner and P.~Blunsom, ``Recurrent continuous translation models,'' in
  \emph{Proceedings of the 2013 Conference on Empirical Methods in Natural
  Language Processing}, 2013, pp. 1700--1709.

\bibitem{cho2014properties}
K.~Cho, B.~Van~Merri{\"e}nboer, D.~Bahdanau, and Y.~Bengio, ``On the properties
  of neural machine translation: Encoder-decoder approaches,'' \emph{arXiv
  preprint arXiv:1409.1259}, 2014.

\bibitem{sut2014seq}
I.~Sutskever, O.~Vinyals, and Q.~V. Le, ``Sequence to sequence learning with
  neural networks,'' in \emph{Advances in neural information processing
  systems}, 2014, pp. 3104--3112.

\bibitem{vas2017atten}
A.~Vaswani, N.~Shazeer, N.~Parmar, J.~Uszkoreit, L.~Jones, A.~N. Gomez,
  {\L}.~Kaiser, and I.~Polosukhin, ``Attention is all you need,'' in
  \emph{Advances in neural information processing systems}, 2017, pp.
  5998--6008.

\bibitem{atret2018unsup}
M.~Artetxe, G.~Labaka, E.~Agirre, and K.~Cho, ``Unsupervised neural machine
  translation,'' \emph{arXiv preprint arXiv:1710.11041}, 2017.

\bibitem{zhang2018style}
Z.~Zhang, S.~Ren, S.~Liu, J.~Wang, P.~Chen, M.~Li, M.~Zhou, and E.~Chen,
  ``Style transfer as unsupervised machine translation,'' \emph{arXiv preprint
  arXiv:1808.07894}, 2018.

\bibitem{gpt2}
A.~Radford, J.~Wu, R.~Child, D.~Luan, D.~Amodei, and I.~Sutskever, ``Language
  models are unsupervised multitask learners,'' \emph{OpenAI Blog}, vol.~1,
  no.~8, p.~9, 2019.

\bibitem{yang2018unsup}
Z.~Yang, Z.~Hu, C.~Dyer, E.~P. Xing, and T.~Berg-Kirkpatrick, ``Unsupervised
  text style transfer using language models as discriminators,'' in
  \emph{Advances in Neural Information Processing Systems}, 2018, pp.
  7287--7298.

\bibitem{li2018style}
J.~Li, R.~Jia, H.~He, and P.~Liang, ``Delete, retrieve, generate: A simple
  approach to sentiment and style transfer,'' \emph{arXiv preprint
  arXiv:1804.06437}, 2018.

\bibitem{nahas2019unsup}
A.~Al~Nahas, M.~S. Tunali, and Y.~S. Akgul, ``Supervised text style transfer
  using neural machine translation: Converting between old and modern turkish
  as an example,'' in \emph{2019 27th Signal Processing and Communications
  Applications Conference (SIU)}.\hskip 1em plus 0.5em minus 0.4em\relax IEEE,
  2019, pp. 1--4.

\bibitem{shang2019semi}
M.~Shang, P.~Li, Z.~Fu, L.~Bing, D.~Zhao, S.~Shi, and R.~Yan, ``Semi-supervised
  text style transfer: Cross projection in latent space,'' \emph{arXiv preprint
  arXiv:1909.11493}, 2019.

\bibitem{jin2019unsup}
Z.~Jin, D.~Jin, J.~Mueller, N.~Matthews, and E.~Santus, ``Unsupervised text
  style transfer via iterative matching and translation,'' \emph{arXiv preprint
  arXiv:1901.11333}, 2019.

\bibitem{luo2019dual}
F.~Luo, P.~Li, J.~Zhou, P.~Yang, B.~Chang, Z.~Sui, and X.~Sun, ``A dual
  reinforcement learning framework for unsupervised text style transfer,''
  \emph{arXiv preprint arXiv:1905.10060}, 2019.

\bibitem{lit2019head}
M.~Litvak, J.~Conroy, and P.~A. Rankel, ``Ranlp 2019 multilingual headline
  generation task overview,'' in \emph{Proceedings of the Workshop MultiLing
  2019: Summarization Across Languages, Genres and Sources}, 2019, pp. 1--5.

\bibitem{li2017rec}
P.~Li, W.~Lam, L.~Bing, and Z.~Wang, ``Deep recurrent generative decoder for
  abstractive text summarization,'' \emph{arXiv preprint arXiv:1708.00625},
  2017.

\bibitem{liu2018gen}
L.~Liu, Y.~Lu, M.~Yang, Q.~Qu, J.~Zhu, and H.~Li, ``Generative adversarial
  network for abstractive text summarization,'' in \emph{Thirty-second AAAI
  conference on artificial intelligence}, 2018.

\bibitem{tan2017abs}
J.~Tan, X.~Wan, and J.~Xiao, ``Abstractive document summarization with a
  graph-based attentional neural model,'' in \emph{Proceedings of the 55th
  Annual Meeting of the Association for Computational Linguistics (Volume 1:
  Long Papers)}, 2017, pp. 1171--1181.

\bibitem{kiy2011opt}
F.~Kiyomarsi and F.~R. Esfahani, ``Optimizing persian text summarization based
  on fuzzy logic approach,'' in \emph{2011 International Conference on
  Intelligent Building and Management}, 2011.

\bibitem{chen2002sum}
F.~Chen, K.~Han, and G.~Chen, ``An approach to sentence-selection-based text
  summarization,'' in \emph{2002 IEEE Region 10 Conference on Computers,
  Communications, Control and Power Engineering. TENCOM'02. Proceedings.},
  vol.~1.\hskip 1em plus 0.5em minus 0.4em\relax IEEE, 2002, pp. 489--493.

\bibitem{san2014sum}
Y.~Sankarasubramaniam, K.~Ramanathan, and S.~Ghosh, ``Text summarization using
  wikipedia,'' \emph{Information Processing \& Management}, vol.~50, no.~3, pp.
  443--461, 2014.

\bibitem{shen2007sum}
D.~Shen, J.-T. Sun, H.~Li, Q.~Yang, and Z.~Chen, ``Document summarization using
  conditional random fields.'' in \emph{IJCAI}, vol.~7, 2007, pp. 2862--2867.

\bibitem{lample2019cross}
A.~Conneau and G.~Lample, ``Cross-lingual language model pretraining,'' in
  \emph{Advances in Neural Information Processing Systems}, 2019, pp.
  7057--7067.

\bibitem{dev2018bert}
J.~Devlin, M.-W. Chang, K.~Lee, and K.~Toutanova, ``Bert: Pre-training of deep
  bidirectional transformers for language understanding,'' \emph{arXiv preprint
  arXiv:1810.04805}, 2018.

\bibitem{artetxe2018unsupMT}
M.~Artetxe, G.~Labaka, E.~Agirre, and K.~Cho, ``Unsupervised neural machine
  translation,'' \emph{arXiv preprint arXiv:1710.11041}, 2017.

\bibitem{kauf2010syn}
M.~Kaufmann and J.~Kalita, ``Syntactic normalization of twitter messages,'' in
  \emph{International conference on natural language processing, Kharagpur,
  India}, 2010.

\bibitem{corro2013clausie}
L.~Del~Corro and R.~Gemulla, ``Clausie: clause-based open information
  extraction,'' in \emph{Proceedings of the 22nd international conference on
  World Wide Web}, 2013, pp. 355--366.

\bibitem{ngu2017rob}
D.~T. Nguyen, K.~A. Al~Mannai, S.~Joty, H.~Sajjad, M.~Imran, and P.~Mitra,
  ``Robust classification of crisis-related data on social networks using
  convolutional neural networks,'' in \emph{Eleventh International AAAI
  Conference on Web and Social Media}, 2017.

\bibitem{clozetask}
W.~L. Taylor, ``“cloze procedure”: A new tool for measuring readability,''
  \emph{Journalism quarterly}, vol.~30, no.~4, pp. 415--433, 1953.

\bibitem{banchs2015adequacy}
R.~E. Banchs, L.~F. D’Haro, and H.~Li, ``Adequacy--fluency metrics:
  Evaluating mt in the continuous space model framework,'' \emph{IEEE/ACM
  Transactions on Audio, Speech, and Language Processing}, vol.~23, no.~3, pp.
  472--482, 2015.

\bibitem{fleiss1971measuring}
J.~L. Fleiss, ``Measuring nominal scale agreement among many raters.''
  \emph{Psychological bulletin}, vol.~76, no.~5, p. 378, 1971.

\bibitem{welch1947generalization}
B.~L. Welch, ``The generalization ofstudent's' problem when several different
  population variances are involved,'' \emph{Biometrika}, vol.~34, no. 1/2, pp.
  28--35, 1947.

\end{thebibliography}

\end{document}